\documentclass[sigconf]{acmart}
\setcopyright{none}
\renewcommand\footnotetextcopyrightpermission[1]{} 

\AtBeginDocument{%
  \providecommand\BibTeX{{%
    \normalfont B\kern-0.5em{\scshape i\kern-0.25em b}\kern-0.8em\TeX}}}

\usepackage{times}
\usepackage{epsfig}
\usepackage{graphicx}
\usepackage{amsmath}

\usepackage{amssymb}
\usepackage{enumitem}

\usepackage{algorithm, algpseudocode}
\usepackage{multirow}
\usepackage{caption}
\usepackage{comment}
\usepackage{amsmath}
\usepackage{amssymb}
\definecolor{citecolor}{RGB}{119,185,0}

\newcommand{\tablestyle}[2]{\setlength{\tabcolsep}{#1}\renewcommand{\arraystretch}{#2}\centering\footnotesize}
\usepackage{subfig}
\usepackage{float}
\usepackage{pifont}
\usepackage{color}
\usepackage{array}

\newlength\savewidth\newcommand\shline{\noalign{\global\savewidth\arrayrulewidth
  \global\arrayrulewidth 1pt}\hline\noalign{\global\arrayrulewidth\savewidth}}

\def\eg{\emph{e.g.}} 
\def\ie{\emph{i.e.}} 
\def\etal{\emph{et~al.}}

\begin{document}

\title{ Towards Unified Text-based Person Retrieval: \\ 
A Large-scale Multi-Attribute and Language Search Benchmark}

\author{  
Shuyu Yang$^1$ \quad Yinan Zhou$^1$ \quad Yaxiong Wang$^{2*}$ \quad Yujiao Wu$^3$ \quad Li Zhu$^{1*}$ \quad Zhedong Zheng$^4$}
\affiliation{
  \institution{$^1$Xi'an Jiaotong University, $^2$Hefei University of Technology, $^3$Pengcheng Lab, $^4$National University of Singapore}
  }
\email{{ysy653, zyn13572297710, wangyx15}@stu.xjtu.edu.cn, yujiaowu111@gmail.com,}
\email{zhuli@mail.xjtu.edu.cn, zdzheng@nus.edu.sg}

\renewcommand{\shortauthors}{Shuyu Yang et al.}

\renewcommand{\thefootnote}{\fnsymbol{footnote}}
\begin{abstract}\footnotetext{$^{*}$Corresponding authors.\vspace{-0.1cm}}
In this paper, we introduce a large Multi-Attribute and Language Search dataset for text-based person retrieval, called MALS, and explore the feasibility of performing pre-training on both attribute recognition and image-text matching tasks in one stone. 
In particular, MALS contains $1,510,330$ image-text pairs, which is about $\textbf{37.5}\times$ larger than prevailing CUHK-PEDES, and all images are annotated with 27 attributes. 
Considering the privacy concerns and annotation costs, we leverage the off-the-shelf diffusion models to generate the dataset.
To verify the feasibility of learning from the generated data, we develop a new joint Attribute Prompt Learning and Text Matching Learning (APTM) framework, considering the shared knowledge between attribute and text.
As the name implies, APTM contains an attribute prompt learning stream and a text matching learning stream. 
(1) The attribute prompt learning leverages the attribute prompts for image-attribute alignment, which enhances the text matching learning.
(2) The text matching learning facilitates the representation learning on fine-grained details, and in turn, boosts the attribute prompt learning.
Extensive experiments validate the effectiveness of the pre-training on MALS, achieving state-of-the-art retrieval performance via APTM on three challenging real-world benchmarks. 
In particular, APTM achieves a consistent improvement of $\textbf{+6.96}\%$, $\textbf{+7.68}\%$, and $\textbf{+16.95}\%$ Recall@1 accuracy on CUHK-PEDES, ICFG-PEDES, and RSTPReid datasets by a clear margin, respectively. 
The dataset, model, and code are available at \url{https://github.com/Shuyu-XJTU/APTM}. 
\end{abstract}


\maketitle

\begin{figure}[t]
\begin{center}
     \includegraphics[width=1\linewidth]{images/examples.pdf}     
\end{center}
\vspace{-.20in}
      \caption{Selected image-text pairs from our MALS (top) and CUHK-PEDES (bottom).
      We could observe that the visual gap between synthetic data and real one is relatively small. 
      In MALS, image-text pairs match almost as well as manual annotation, although there are some flaws occasionally. 
      It is worth noting that images in MALS are high-fidelity with rich and diverse variations in terms of pose, appearance, background, \emph{etc}. (Best viewed when zooming in.)
      }\label{fig:examples}
      \vspace{-.20in}
\end{figure}

\section{Introduction}

Given the pedestrian description, text-based person retrieval aims to locate the person of interest from a large pool of candidates~\cite{li2017identity}. 
Compared to conventional image-based person retrieval~\cite{zheng2011person, xiao2016end, liao2015person}, text-based person retrieval provides an intuitive way to form queries. 
Such techniques can be widely applied to promote public safety, such as locating lost children in large areas like airports. 
However, as a type of cross-modal learning task, text-based person retrieval harvests little benefits from large-scale cross-modal pretraining. 
The reasons stem from two aspects: 
1) \textbf{Lack of Data.} Due to privacy concerns, we usually can not collect enough data for the current data-hungry deeply-learned models. 
2) \textbf{Lack of High-quality Annotation.} The language annotation process is also tedious and inevitably introduces annotator biases.
As a result, the sentences are usually quite short, which can not comprehensively describe the characteristic of the target person~\cite{li2017person, ding2021semantically}.

In response to these problems, we propose to construct a synthetic image-text dataset, borrowing the power of the off-the-shelf diffusion models and the image caption model. 
In this way, we could generate unlimited images and acquire high-quality annotations automatically.
Furthermore, to make the synthetic data beneficial for real-world language-based person retrieval, there are still two challenges that need to be addressed: 
\textbf{(1) Realism of synthetic image-text pairs.} 
The visual disparity between synthetic and real-world image-text pairs constitutes a major challenge in the construction of a meaningful text-pedestrian benchmark.
For text inputs, we utilize descriptions derived from real-world text-based person data to guide the diffusion models. 
Therefore, the generated images closely resemble those found in the real world.
We further apply a post-processing mechanism as a supplementary step (\emph{See Section~\ref{sec: postprocessing}.}) to further refine the synthetic images and rectify any remaining discrepancies.
\textbf{(2) Diversity of annotations (sentences \& attributes).} 
To generate a large-scale cross-modal dataset, the human-annotated description will inevitably be used multiple times, resulting in poor text diversity. 
To handle this limitation, we employ an off-the-shelf caption generation model to augment the descriptions for each synthetic image. 
Besides, we propose an automatic attribute extraction mechanism that mines the key attributes from the descriptions to further enrich the annotations.

In this way, we collect a new large-scale cross-modal dataset, \emph{i.e., } \textbf{M}ulti-\textbf{A}ttribute and \textbf{L}anguage \textbf{S}earch dataset for person retrieval (MALS) with rich annotations. 
It is worth noting that while diffusion models have been recently studied for data augmentation \cite{azizi2023synthetic, shiparddiversity, sariyildiz2023fake}, 
these works mainly focus on coarse-grained category recognition benchmarks such as ImageNet \cite{5206848} and EuroSAT \cite{helber2019eurosat}.
Differently, person retrieval requires a more detailed representation since the variations among individuals are comparatively small. 
Therefore, the MALS dataset focuses on providing fine-grained details, which is crucial for text-based person retrieval tasks. 
Furthermore, extensive experiments verify that the knowledge learned from MALS is also scalable to real-world applications in terms of both text-based person retrieval and pedestrian attribute recognition tasks.

To verify the value of the collected dataset, 
we introduce an \textbf{A}ttribute \textbf{P}rompt Learning and \textbf{T}ext \textbf{M}atching Learning (APTM) framework for text-based person retrieval. 
As shown in Figure~\ref{fig: framework}, the proposed APTM comprises three modules, the image encoder, text encoder, and cross encoder. 
We utilize text to acquire attribute annotation by the proposed Explicit Matching (EM) and Implicit Extension (IE) mechanism, and further map attributes to a set of Attribute Prompts. 
Image-text contrastive learning (ITC) and image-attribute contrastive learning (IAC) act on the embeddings of feature encoders, while image-text matching (ITM), image-attribute matching (IAM), masked language modeling (MLM), and masked attribute prompt modeling (MAM) are imposed on the respective predictions from the cross encoder. 
The above constraints are jointly optimized during pre-training to learn an effective model.
In summary, we highlight the contributions of this paper as follows:
\begin{itemize}[leftmargin=*]
    \item We observe that data scarcity largely compromises text-based person retrieval. Therefore, we introduce a new large-scale multi-attribute and language search benchmark, called MALS. Compared with the existing datasets, such as CUHK-PEDES, our benchmark contains about $\textbf{37.5}\times$ images with rich attribute annotations.
    (\emph{See Table~\ref{table:Dataset}.})
    
    \item Based on MALS, we also introduce a new joint Attribute Prompt Learning and Text Matching Learning (APTM) framework, to facilitate the representation learning.
    As the name implies, we explicitly leverage both the attribute recognition task and the text-based person retrieval task to regularize the model training. The two tasks are complementary and benefit each other.
    
    \item The proposed approach achieves a competitive recall rate on three challenging real-world benchmarks including CUHK-PEDES, ICFG-PEDES, and RSTPReid. 
    Besides, we observe that the text matching task facilitates attribute recognition as well.
    Fine-tuning APTM on PA-100K, \ie, a prevalent pedestrian attribute recognition dataset, we obtain competitive performance $82.58 \%$ mA. 
\end{itemize}

\section{Related Work}

\textbf{Language-based Person Search.}
Text-to-image person retrieval is more challenging than general cross-modal retrieval tasks because of its fine-grained nature. 
Existing efforts can be classified as cross-modal attention-based~\cite{li2017person, wang2022look, shu2023see, shao2022learning} approaches or cross-modal attention-free approaches~\cite{ding2021semantically, zheng2020dual, wang2022caibc, chen2022tipcb} depending on the alignment strategy.
To align representations from both modalities in a shared feature space, 
the cross-modal attention-free approaches build various model structures or objective functions~\cite{zheng2020dual}. 
In contrast, cross-modal attention-based approaches require pair-wise inputs and
encourage building cross-modal correspondences between regions and words or regions and phrases with more interactions between modalities. 
It is worth noting that both strategies have their advantages as well as disadvantages. 
In general, cross-modal attention-free techniques are more efficient. More specifically, their complexity is $O(M+N)$ for $M$ gallery and $N$ queries. 
The complexity of cross-modal attention-based approaches, in comparison, rises to $O(MN)$ due to the pair-wise inputs. 
Nonetheless, these techniques typically result in noticeably superior retrieval performance. 
It is because cross-modal attention-based approaches reduce modality gaps more effectively with more cross-modality communication in an early stage.
In this paper, we leverage cross-modal attention-free features to quickly find the candidates and then deploy the attention-based module to refine the final ranking score.

\noindent\textbf{Attribute-based Person Re-identification.}
Attribute-based person re-identification~\cite{lin2019improving, han2018attribute, ling2019improving, luo2019learning, zhang2019part, tay2019aanet, li2020attributes} aims to identify individuals across different cameras or time periods based on their attributes, such as clothing color, gender, height, \emph{etc.}, rather than relying solely on visual appearance.
One of the earliest works on pedestrian attributes is by Lin \etal~\cite{lin2019improving}, who propose a framework for person re-identification using color, texture, and contour clues. In particular, Lin \etal~ extract discriminative features from each pedestrian image and train several attribute classifiers. 
Following this work, Han~\etal~\cite{han2018attribute} further propose to fuse part features with attribute attentions, while He~\etal~\cite{he2017adaptively} study to jointly train multiple attribute classifiers in a coherent manner. 
In contrast to the fixed horizontal splitting, attribute localization is also studied in~\cite{shi2020person}.
To encourage the interaction between attributes, both Nguyen \etal~\cite{nguyen2021graph} and Tang \etal~\cite{tang2022learning} propose to build a graph representation of the attributes for each person, where nodes represent attribute embeddings and edges represent correlations between them. 
In addition to traditional attributes, Wang \etal~\cite{wang2018transferable} leverages both appearance and personality traits to learn representations of both visual appearance and personality traits and combine them for re-identification. 
Finally, there are several works that make attributes more robust against occlusions or pose variations. 
For instance, Jing \etal~\cite{jing2020pose} propose a multi-modal framework that fuses attribute-based features with pose-based features to enhance re-identification accuracy under challenging conditions. 
In this paper, we also leverage robust attribute learning to facilitate text-based person retrieval. We find that attribute learning is complementary to image-text matching, and vice versa.

\setlength{\tabcolsep}{5pt}
\vspace{-.2in}
\begin{table*}[t]
\small
\begin{center}
\begin{tabular}{l|c|c|c|c|c|c}
\hline
Datasets & MALS (Ours) & CUHK-PEDES~\cite{li2017person} & ICFG-PEDES~\cite{ding2021semantically} & RSTPReid~\cite{zhu2021dssl} & Lin \etal~\cite{lin2019improving} & PA-100K~\cite{liu2017hydraplus} \\
\shline
\#Images           &  \textbf{1,510,330}  &  40,206  &  54,522  &  20,505  &  32,668  &  100,000  \\

\multirow{2}{*}{Data Source} & Automatic & Market~\cite{Zheng_2015_ICCV} \& & \multirow{2}{*}{MSMT-17~\cite{wei2018person}} & \multirow{2}{*}{MSMT-17~\cite{wei2018person}} & Market~\cite{Zheng_2015_ICCV} \& &  Manual \\
 & Synthesis & Duke~\cite{ristani2016performance}, \emph{etc.} & & & Duke~\cite{ristani2016performance} &  Collection \\
\#Avg Texts/Image  &  1  &  2  &  1  &  2  &  -  &  -  \\
\#Avg Text Length  &  26.96  &  23.54  &  37.2  &  25.8  &  -  &  -   \\ 
Surrounding        &  Indoor \& Outdoor  &  Indoor \& Outdoor  &  Indoor \& Outdoor  &  Indoor \& Outdoor  &  Outdoor  &  Outdoor  \\
Resolution         &  $531\times 208$  &  $246\times 90$  &  $378\times 142$  &  $546\times 219$  &  $128\times 64$  &  $225\times 85$  \\
Annotation         &  \textbf{Sentence \& Attribute}  &  Sentence  &  Sentence  &  Sentence  &  Attribute  &  Attribute  \\
\#Attribute        &  \textbf{27}  &  -  &  -  &  -  &  \textbf{27}  &  26  \\
\hline
\end{tabular}
\end{center}
\caption{Comparison between MALS and other real-world datasets for text-based person retrieval and pedestrian attribute recognition. Current datasets typically collect images from existing person re-ID datasets and manually provide corresponding natural language descriptions or attribute annotations. In contrast, MALS leverages generative models to generate a large-scale dataset including $1.5M$ image-text pairs.
For each benchmark, the table shows the number of images, data source, the average texts per image, average text length, the main surrounding and the average resolution of images, types of annotations as well as the number of attributes.
}
\label{table:Dataset}
\vspace{-.3in}
\end{table*}

\setlength{\tabcolsep}{3pt}
\begin{table}
\small
\begin{center}
\resizebox{1\linewidth}{!}{\begin{tabular}{l|c|c}
\hline
Attribute Category  & Name   &  Label  \\
\shline
gender              &  gender  &  female(0), male(1)  \\
age	                &  age  &  young(0), teenager(1), adult(2), old(3)  \\
length hair         &  hair  &  short hair(0), long hair(1)  \\
wearing hat         &  hat  &  yes(0), no(1)  \\
carrying backpack   &  backpack  &  yes(0), no(1)  \\
carrying handbag	&  handbag  &  yes(0), no(1)  \\
carrying bag        &  bag  &  yes(0), no(1)  \\
sleeve length       &  sleeve  &  long sleeve(0), short sleeve(1)  \\
length of lower-body  &  length\_lower & long lower body clothing(0), short(1)  \\
type of lower-body    &  type\_lower & dress(0), pants(1)  \\
\multirow{2}{*}{color of upper-body}   &  black, white, red, purple,   & (0), (1), (2), (3),   \\
& yellow,blue, green, gray & (4),(5), (6), (7) \\
\multirow{2}{*}{color of lower-body}   &  black, white, purple, yellow, & (0), (1), (2), (3),    \\
& blue, green, pink, gray, brown & (4),(5), (6), (7), (8)\\
\hline
\end{tabular}}
\end{center}
\caption{Attribute space consists of $27$ attributes.
Here we show the attribute category, the name in the annotation file, and the available label choices.
}
\label{table:Attribute}
\vspace{-.3in}
\end{table}

\section{Benchmark} \label{sec:dataset}

Existing text-based person retrieval datasets~\cite{li2017person, ding2021semantically, zhu2021dssl} typically collect pedestrian images from existing person re-identification datasets and manually annotate corresponding text descriptions. However, such practice greatly limits the scale and diversity due to annotation costs and privacy concerns, as shown in Table~\ref{table:Dataset}.
The great success of recent diffusion models ~\cite{rombach2022high,brooks2022instructpix2pix, hertz2022prompt} inspires us to collect pedestrian images from the synthetic domain. 
There are two primary advantages: 
(1) Comparing to 3D Game Engine~\cite{sun2019dissecting,xiang2021less,wang2020surpassing} or Generative Adversarial Networks (GANs)~\cite{zheng2017unlabeled,tang2019cycle,zheng2019joint,jiang2021exploring}, diffusion models have shown a strong and stable ability to synthesize images with high authenticity to text, significantly reducing the gap between synthetic and real data. (2) Using synthetic pedestrian images also circumvents privacy concerns. The construction of our benchmark consists of the following steps:

\noindent{\textbf{Image-text Pair Generation.}} 
We utilize the off-the-shelf diffusion model, ImaginAIry \cite{imaginAIry} which could generate new pedestrian images. 
To make the generated samples reasonable as well as close to the real-world pedestrian images, we employ the textual descriptions of the CUHK-PEDES~\cite{li2017person} dataset and the ICFG-PEDES~\cite{ding2021semantically} dataset as prompts. 
We feed the prompts into ImaginAIry and collect the corresponding synthetic images, resulting in a pair of aligned samples. To ensure the generation of high-quality full-body pedestrian images with controlled variability, we set the image size as $576 \times 384$ and adjust the random seed to get the high-quality samples. By randomizing the noise during inference, massive and diverse pedestrian images are collected.

\label{sec: postprocessing}
\noindent{\textbf{Post-Processing.}} 
Due to the lack of fine-grained and controllable generation capabilities of the text-to-image generation model, many generated images cannot meet the requirement of training the pedestrian retrieval networks. Two main issues stand: (1) the low-quality images, including grayscale and blur images. 
To overcome this weakness, we simply sort images by file size and delete images whose size is smaller than $24k$ to filter out blurred images. Then we compute the mean variance of the difference between the 3 channels of every image and remove images whose mean variance is less than a presetting threshold.
(2) the noisy images, \eg, multiple persons in one image, only part of a person, and no person. To remedy this issue, we apply OpenPose \cite{8765346, simon2017hand, cao2017realtime, wei2016cpm} to detect human key points and filter out the undesired person images. We also leverage the detected key points as a tight bounding box to re-crop the samples. With the above steps, we acquire the final pedestrian images.

\begin{figure*}[t]
\vspace{-.1in}
\begin{center}
     \includegraphics[width=1\linewidth]{images/tagcloud.pdf}     
\end{center}
\vspace{-.20in}
      \caption{Two tag clouds based on the texts from MALS  and CUHK-PEDES~\cite{li2017identity}, separately. We could observe that text descriptions from MALS and CUHK-PEDES share a significant amount of common corpus despite domain differences. The sharing content facilitates the transfer of pre-trained models from large-scale synthetic data to real-world data.
      }\label{fig:tag}
      \vspace{-.1in}
\end{figure*}

\noindent{\textbf{Caption Calibration.}}
The prompts used to generate images are the straightforward choice to serve as the text descriptions.
However, this fashion would result in poor diversity of the textural descriptions, since multiple images usually share the same text. 
To cope with this problem, we leverage the cross-modal model, BLIP \cite{li2022blip} to produce more fitting captions for every synthetic image and form the final image-text pairs.

\vspace{-0.08cm}
\noindent{\textbf{Attribute Annotation.}} 
The associated attributes often highlight the key characteristics of both image and text samples, and many works of text-based person retrieval indicate the potential of attribute for performance improvement~\cite{shu2023see, ding2021semantically, chen2022tipcb}. Inspired by this, we further augment our MALS with the attribute annotation, so that a more informative and comprehensive benchmark can be constructed. Considering the cost of manual annotation, we obtain the attribute annotation in an automatic manner.
We first define the attribute space in the same way as Market-1501 Attribute~\cite{lin2019improving}, and then propose two mechanisms to obtain attributes, Explicit Matching (EM) and Implicit Extension (IE). 
EM deploys the correspondence of specific attributes based on keywords in the text, such as the word "man" corresponding to the attribute "gender: male". 
IE assigns corresponding attribute candidates based on distinctive features that are not mentioned in the text, such as allocating samples that do not mention "hat" in their descriptions to the attribute "hat: no". 
Finally, $27$ different types of attributes are collected, as shown in Table \ref{table:Attribute}.

\noindent{\textbf{MALS Benchmark.}}
Following the above steps, a high-fidelity, diverse and large-scale benchmark for the text-based person retrieval task is built. 
As shown in Figure~\ref{fig:examples}, we can observe the quality of visual images and textual sentences are comparable with CUHK-PEDES.
Figure~\ref{fig:tag} also intuitively presents a comparison of the word distributions of our MALS and CUHK-PEDES using word clouds. We could observe that, although there still exist several differences between the two datasets, the text corpus of MALS is close to the real-world data.
Compared with existing text-based person retrieval datasets in Table~\ref{table:Dataset}, MALS has the following advantages:

\begin{itemize}[leftmargin=*]
\item \textbf{High-fidelity Images:} Compared with the images with poor lighting and blur texture, collected from surveillance cameras, images of MALS are of higher quality benefiting from the ability of the diffusion model (\emph{see Figure~\ref{fig:examples}.}), which means that the synthetic images are more visually appealing and realistic.

\item \textbf{Diversity:} MALS contains a wide range of variations in the images, including but not limited to variations in background, viewpoint, occlusion, clothing, and body pose. Thanks to our caption calibration step, the associated textual descriptions are also diverse enough. Therefore, MALS can support us to train robust models that generalize well to new and unseen data in vision tasks, language tasks, and vision-language tasks.

\item \textbf{Fewer Privacy Concerns:} Unlike several traditional benchmarks of text-based person retrieval capturing images without consent, the samples of our MALS are all synthetic images generated by the off-the-self stable diffusion model, which avoids ethical and legal issues.

\item \textbf{Large-scale Pairs:} MALS contains $1.5M$ image-text pairs (\emph{see Table \ref{table:Dataset}.}), while existing datasets usually provide no more than $100k$ of aligned image-text. This magnitude of the dataset enables a comprehensive pre-training study.

\item \textbf{Rich Annotations:} Each image-text pair in MALS is annotated with appropriate attribute labels, indicating that MALS is not only effective for text-image matching and attribute prompt learning, but also explores the feasibility of pre-training for both attribute recognition and image-text matching in one stone.
\end{itemize}

\begin{figure*}[t]
\vspace{-.1in}
\begin{center}
     \includegraphics[width=0.95\linewidth]{images/framework.pdf}
\end{center}
\vspace{-.2in}
      \caption{Overview of the proposed Attribute Prompt Learning and Text Matching Learning (APTM) framework for pre-training on MALS. APTM framework contains one image-attribute stream and one image-text stream with weight-shared encoders. In particular, the framework comprises three encoders, \ie, Image Encoder ($E_I$), Text Encoder ($E_T$), Cross Encoder ($E_C$), and two MLPs-based headers.  The Image Encoder and Text Encoder are to produce the embeddings of the image and text, respectively, while the cross encoder seeks to fuse the image and text embeddings for the subsequent predictions.}\label{fig: framework}
      \vspace{-.1in}
\end{figure*}

\section{Method} \label{method}
We leverage MALS as a pre-training dataset and devise a new simple joint Attribute Prompt Learning and Text Matching Learning (APTM) framework, as shown in Figure~\ref{fig: framework}. The overall pipeline is typically divided into two steps, \emph{i.e.,} pre-training and fine-tuning.  
During pre-training, we perform Attribute Prompt Learning (APL) and Text Matching Learning (TML) to learn the common knowledge of text-based person retrieval and pedestrian attribute recognition. In the second step, the parameters are further optimized toward a specific downstream task.
In this section, we elaborate on the details of the pre-training stage, as we mainly study the benefits of our MALS for pre-training.

\subsection{APTM Architecture} 
As shown in Figure~\ref{fig: framework}, APTM is a multi-task framework, containing one image-attribute stream and one image-text stream with weight-shared encoders and MLP-based headers. 
In particular, the framework comprises three encoders, \ie, Image Encoder ($E_I$), Text Encoder ($E_T$), Cross Encoder ($E_C$), and two MLPs-based headers.  
Before pre-training, we utilize text to acquire attribute annotation by Explicit Matching and Implicit Extension mechanism and then map attributes to a set of Attribute Prompts as one of the inputs of the image-attribute stream. 
During pre-training, the image-text stream and the image-attribute stream are jointly trained. We deploy Random Mask to generate masked text and masked attribute prompts, and then the Image Encoder maps the image into embedding $V$ and the Text Encoder extracts different text representations by encoding Text, Masked Text, Attribute Prompts, and Masked Attribute Prompts separately, denoted as $L$, $\hat{L}$, $L_A$ and $\hat{L_A}$, respectively. 
In the task of ITC and ITM, $V$ is paired with $L$, while in the context of IAC and IAM, $V$ is paired with $L_A$. Further, $V$ is also fed into cross encoder with $\hat{L}$ or $\hat{L_A}$ for MLM or MAM task.

\noindent\textbf{Image Encoder.} 
Without loss of generality, we deploy Swin Transformer (Swin-B)~\cite{liu2021swintransformer} as Image Encoder ($E_I$).
Given an image ($I$) of resolution of $384 \times 128$, we split it into $N^I$ non-overlapping patches with a patch size of $32 \times 32$, where $N^I = 48$. 
Then, these patches are linearly embedded and passed into the transformer layers of $E_I$, yielding a set of high-dimensional embeddings $V$, the \texttt{[CLS]} embedding $v^{cls}$ is taken as the representation of the entire image.

\noindent\textbf{Text Encoder.}
Following existing works~\cite{shu2023see}, we intuitively employ BERT~\cite{kenton2019bert} as Text Encoder ($E_T$) for a fair comparison. Specifically, the text ($T$) is first tokenized as ${N^T + 1}$ tokens and fed into the first 6 layers of BERT. The output text embeddings $\{l^{cls}, l^1, l^2, ..., l^{N^T}\}$ is denoted as $L$, where ${l^i (i \in [1, N^T])}$ represents the embedding of the ${i_{th}}$ text token. The embedding of the \texttt{[CLS]} token, \ie,~${l^{cls}}$ is treated as the whole text representation.

\noindent\textbf{Cross Encoder.}
The cross encoder is to fuse the image and text representations to perform the prediction tasks. Specifically, we adopt the last 6 layers of BERT as Cross Encoder ($E_C$). As shown in Figure~\ref{fig: framework}, the image and text embeddings are fed into $E_C$ and fused by the cross attention mechanism to capture their semantic relationship. Finally, the joint representation can be obtained: $C = \{c^{cls}, c^1, c^2, ..., c^{N^T}\}$.

\subsection{Attribute Prompt Learning }

\textbf{Motivations.}
Attributes often emphasize crucial characteristics of pedestrian images, such as gender and hair, which are vital for performing cross-modal alignment and distinguishing between candidates. 
Moreover, as depicted in Figure~\ref{fig:tag}, synthetic and real descriptions exhibit a considerable overlap in attribute keywords, leading us to believe that accentuating the similar attribute space can also alleviate the domain gap. 
To better leverage the attribute information for image-attribute alignment, we have opted not to rely on conventional classifier-based multi-attribute learning methods.
Instead, we convert attribute labels into attribute prompts with prompt templates, as illustrated in Figure~\ref{fig: framework}. 
We then align the attribute prompts with the corresponding image, which forms the fundamental basis of our Attribute Prompt Learning.
Drawing inspiration from cross-modal Learning, we utilize Image-Attribute Contrastive Learning (IAC), Image-Attribute Matching (IAM), and Masked Attribute Language Modeling (MAM) to effectively align images with their attributes.

\noindent\textbf{Image-Attribute Contrastive Learning} 
(IAC) concentrates on mastering the ability to differentiate between positive and negative pairs. 
Given a set of attribute texts $\{T_a^k\}^{2|A|}$ in a mini-batch, $k \in [1, 2|A|]$, where $A$ is the attribute set of $27$ binary attributes.
For an image $I$, if any of its attribute labels correspond with the attribute set, we consider the corresponding attribute text and $I$ as a matched (image, attribute prompt) pair. If not, they are considered unmatched. 
As exemplified in Figure~\ref{fig: framework}, "the person is a man" is a matched attribute prompt of the image while "the person is a woman" is not. We denote the set of all matched (image, attribute prompt) pairs in a mini-batch as $B_a$.
The matching score between an image $I$ and its paired attribute prompt $T_a$ is estimated as follows:
\vspace{-.1in}
\begin{equation}
\label{itoa}
S_{\text{i2a}}(I) = \frac{\exp(s(F_I, F_{T_a})/\tau)}{\exp(s(F_I, F_{T_a})/\tau) + \exp(s(F_I, F_{\bar{T}_a})/\tau)},
\end{equation}
where ${\bar{T}_a}$ is the opposite attribute prompt of ${T_a}$, {which is constructed by replacing the true attribute as the false one, \emph{e.g., man$\Rightarrow$woman}}, $\tau$ is a learnable temperature parameter, $F_I$ and $F_{T_a}$ are the mapped features of their respective \texttt{[CLS]} embedding by two different FCs, $s(\cdot,\cdot)$ is the cosine similarity. Finally,
the formulation of the IAC loss is presented below:
\vspace{-.1in}
\begin{equation}
\begin{split}
\mathcal{L}_{iac} = -\frac{1}{|B_a|}\sum_{(I,T_a) \in B_a}\log S_{\text{i2a}}(I).
\end{split}
\end{equation}

\noindent\textbf{Image-Attribute Matching Learning}
(IAM) aims to predict whether the input image and attribute prompt are matched. 
In particular, IAM is specified as a binary classification problem to facilitate the image-attribute alignment: the positive sample is the paired image-attribute prompt, while the unpaired is the negative one.
Mathematically, assume $|B|$ images are sampled in a mini-batch, $5$ attribute prompts are randomly constructed to form $5|B|$ (image, attribute prompt) pairs, denote as $\bar{B}_a$.
Subsequently, the image-attribute prompt tuples are passed through the Cross Encoder to get the \texttt{[CLS]} embedding $c^{cls}$, their matching score is given by an MLP with Sigmoid activation: $p^{\text{match}}(I, T_a) = \text{Sigmoid}(\text{MLP}(c^{cls}))$,
the IAM loss is defined as: 
\vspace{-.1in}
\begin{equation}
\begin{split}
\mathcal{L}_{\text{iam}} =& -\frac{1}{|\bar{B}_a|} \sum_{(I, T_a)\in \bar{B}_a}(y^{\text{match}}_a\log p^{\text{match}}(I, T_a)\\
&+ (1-y^{\text{match}}_a)(1-\log p^{\text{match}}(I, T_a))), \label{eq:litm} 
\end{split}
\end{equation}
where $y^{\text{match}}_a$ is 1 if $(I, T_a)$ is matched, 0  otherwise.

\noindent\textbf{Masked Attribute Language Modeling}
(MAM) seeks to predict the masked words using the matched (image, attribute prompt) as a clue.
To this end, we first adopt the following strategies to randomly mask the $2|A|$ attribute prompts:
1) mask out the text tokens with a probability of $25\%$; 
Among the masked tokens, 2) $10\%$ and $80\%$ is replaced with random tokens and the special token \texttt{[MASK]}, respectively; 
3) $10\%$ remain unchanged.
Then, given an image-attribute prompt pair $(I, T_a)$ in $B_a$, we obtain corresponding masked attribute prompt $\hat{T}_a$ following the aforementioned strategies.
Then, $(I, \hat{T}_a)$ is input into encoders to get the output of $E_C$: $\hat{C} = \{\hat{c}^{cls}, \hat{c}^1, \hat{c}^2, ..., \hat{c}^{N^T}\}$.
If $\hat{t}^j_a$ is the masked token in $\hat{T}_a$, $j \in [1, N^T]$, its prediction probability is given by an MLP with Softmax activation: $p^{\text{mask}}_j(I, \hat{T}_a) = \text{Softmax}(\text{MLP}(\hat{c}^{j}))$. Finally, the MAM loss is defined as follows:
\vspace{-.1in}
\begin{equation}
\label{malmLoss}
\mathcal{L}_{\text{mam}} = \mathbb{E}_{\hat{t}^j_a \sim \hat{T}_a;(I,\hat{T}_a) \sim \hat{B}_a}H(y^{\text{mask}}_j, p^{\text{mask}}_j(I, \hat{T}_a)), 
\end{equation}
where $y^{\text{mask}}_j$ is a one-hot distribution in which the ground-truth
token $\hat{t}^j_a$ has the probability of one, and $\hat{B}_a$ is the $2|A|$ (image, masked attribute prompt) pairs of the mini-batch.
The overall APL loss is: $\mathcal{L}_{\text{APL}} = \frac{1}{3}(\mathcal{L}_{\text{iac}} + \mathcal{L}_{\text{iam}} +\mathcal{L}_{\text{mam}}).$
To prevent overfitting, label smoothing is further employed. Typically, we apply a random noise perturbation to $y^{\text{match}}_{a}$, which remedies overconfident predictions.

\noindent\textbf{Why APL Works Better.} 
In comparison to the Classification-based Multi-Attribute Learning (CMAL) approaches, APL has three distinct advantages: 
1) Explicit emphasis on the attributions. Na\"ive classification-based practices implicitly highlight the key attributes through a classification procedure, whereas APL explicitly constructs the attribute prompt, which leads to more effective learning than implicit classification procedures.
2) More informative inputs. APL introduces information-rich inputs by constructing supplementary attribute prompts, providing richer information for cross-modal alignment learning. In contrast, traditional CMAL only utilizes a classification loss and introduces no auxiliary information.
3) Greater flexibility for framework augmentation. Thanks to the constructed attribute prompts, APL enables powerful cross-modal learning objectives such as image-text contrastive learning (ITC), Image-text matching (ITM), and masked language modeling (MLM) to be equipped after modification to be attribute-oriented, resulting in increased potential for performance improvement. During experiments, APL outperforms several na\"ive CMAL variants, which well verifies the superiority of APL.

\subsection{Text matching Learning} 
As a type of cross-modal retrieval problem, the core of text-based person retrieval is to align the text query and the image candidates. Hence, we also incorporate the tasks of Image-Text Contrastive Learning (ITC), Image-Text Matching Learning (ITM), and Masked Language Modeling (MLM) to impose the alignment constraints.

\noindent\textbf{Image-Text Contrastive Learning}
(ITC) focuses on learning to differentiate between positive and negative pairs. In our case, it is intuitive to treat the paired image-text (I, T) as the positive sample, while the unmatched image-text is the negative pair. 
Formally, we randomly sample $|B|$ pairs of images and text in each mini-batch. Similar to Eq.~\ref{itoa}, given a matched pair $(I, T)$, we initially extract their respective representations $F_I$ and $F_T$. The matching score is then estimated as follows:
\vspace{-.1in}
\begin{equation}
S_{\text{i2t}}(I) = \frac{\exp(s(F_I, F_T)/\tau)}{\sum_{i=1}^{|B|}\exp(s(F_I, F_{T^i})/\tau)}, \label{eq:si2t}
\end{equation}
Similarly, given the text, the matching score of the paired image  $S_{\text{t2i}}(T)$ can be calculated.
Finally, the ITC loss is formulated as:
\vspace{-.1in}
\begin{equation}
\begin{split}
\mathcal{L}_{\text{itc}} = -\frac{1}{2|B|}\sum_{(I,T)\in B}(\log S_{\text{i2t}}(I) + \log S_{\text{t2i}}(T)), \label{eq:litc}
\end{split}
\end{equation}
where $B$ is the data set of the mini-batch.

\begin{figure*}
\centering
\begin{minipage}[t]{0.45\linewidth}
\vspace{-1.3cm}
    \centering
    \small    \makeatletter\def\@captype{table}\makeatother
    \centering
    \resizebox{1\linewidth}{!}{ 
        \renewcommand\arraystretch{1} 
        \begin{tabular}[h!]{p{2.6cm}|m{1.2cm}<{\centering}m{1.2cm}<{\centering}m{1.2cm}<{\centering}m{1.2cm}<{\centering}}
            \shline
            \textbf{Method} & \textbf{R1} & \textbf{R5} & \textbf{R10} & \textbf{mAP}  \\
            \hline
            CNN-RNN~\cite{reed2016learning} & 8.07 & - & 32.47 & - \\ 
            GNA-RNN~\cite{li2017person} & 19.05 & - & 53.64 & - \\ 
            PWM-ATH~\cite{chen2018} & 27.14 & 49.45 & 61.02 & - \\ 
            GLA~\cite{chen2018improving} & 43.58 & 66.93 & 76.2 & - \\
            Dual Path~\cite{zheng2020dual} & 44.40 & 66.26 & 75.07 & - \\
            CMPM+CMPC~\cite{zhang2018deep} & 49.37 & - & 79.21 & - \\
            MIA~\cite{niu2020improving} & 53.10 & 75.00 & 82.90 & -\\
            A-GANet~\cite{liu2019deep} & 53.14 & 74.03 & 81.95 & - \\  
            ViTAA~\cite{wang2020vitaa} & 55.97 & 75.84 & 83.52 & 51.60 \\ 
            IMG-Net~\cite{wang2020img} & 56.48 & 76.89 & 85.01 & - \\
            CMAAM~\cite{aggarwal2020text} & 56.68 & 77.18 &	84.86 & - \\
            HGAN~\cite{zheng2020hierarchical} &	59.00 & 79.49 & 86.62 & - \\
            NAFS~\cite{gao2021contextual} &	59.94 & 79.86 & 86.70 & 54.07 \\
            DSSL~\cite{zhu2021dssl} & 59.98 & 80.41 & 87.56 & - \\
            MGEL~\cite{wang2021text} & 60.27 & 80.01 & 86.74 & - \\
            SSAN~\cite{ding2021semantically} & 61.37 & 80.15 & 86.73 & - \\
            NAFS~\cite{gao2021contextual} & 61.50 & 81.19 & 87.51 & - \\
            TBPS~\cite{han2021text} & 61.65 &80.98 & 86.78 & - \\
            TIPCB~\cite{chen2022tipcb} & 63.63 &82.82 & 89.01 & - \\
            LBUL~\cite{wang2022look} & 64.04 &82.66 & 87.22 & - \\
            CAIBC~\cite{wang2022caibc} & 64.43 & 82.87 &88.37 & - \\
            AXM-Net~\cite{farooq2022axm} & 64.44 & 80.52 & 86.77 & 58.73 \\
            SRCF~\cite{suo2022simple} & 64.88 &83.02 & 88.56 & - \\
            LGUR~\cite{shao2022learning} & 65.25 & 83.12 & 89.00 & - \\
            CFine~\cite{yan2022clip} & 69.57 & 85.93 & 91.15 & - \\
            \hline
            TextReid (RN50)~\cite{han2021text} & 61.65 & 80.98 & 86.78 & 58.29 \\
            TextReid (30K)~\cite{han2021text} & 61.81 & 80.39 & 86.90 & - \\
            IVT (Baseline)~\cite{shu2023see} & 55.75 & 75.68 & 84.13 & - \\
            IVT (4M)~\cite{shu2023see} & 65.59 & 83.11 & 89.21 & - \\
            IVT (MALS)~\cite{shu2023see} & 66.10 & 83.79 & 89.46 & - \\
            \hline
            Baseline & 66.44 & 84.92 & 90.76 & 59.19\\
            APTM (Ours) &\textbf{76.53}	& \textbf{90.04} & \textbf{94.15} & \textbf{66.91}\\
            \shline
        \end{tabular}
    }
    \vspace{-0.25cm}
    \caption{Performance Comparison on CUHK-PEDES.}
    \label{tab:sota_CUHK}
\end{minipage}
\hspace{0.4cm}
\quad
\begin{minipage}[t]{0.45\linewidth}
    \centering
    \begin{minipage}[t]{1.0\linewidth}
        \centering
        \small
        \makeatletter\def\@captype{table}\makeatother
        \centering
        \resizebox{1\linewidth}{!}{
            \renewcommand\arraystretch{1} 
            \begin{tabular}[h!]{p{3cm}|m{1.2cm}<{\centering}m{1.2cm}<{\centering}m{1.2cm}<{\centering}m{1.2cm}<{\centering}} 
                \shline
                \textbf{Method} & \textbf{R1} & \textbf{R5} & \textbf{R10} & \textbf{mAP}  \\
                \hline
                DSSL~\cite{zhu2021dssl} & 32.43 & 55.08 & 63.19 & - \\
                LBUL~\cite{wang2022look} & 45.55 & 68.20 & 77.85 & - \\
                IVT~\cite{shu2023see} & 46.70 & 70.00 &	78.80 & - \\
                CAIBC~\cite{wang2022caibc} & 47.35 & 69.55  & 79.00 & - \\
                CFine~\cite{yan2022clip} & 50.55 & 72.50 & 81.60 & - \\
                \hline
                Baseline & 47.20 & 70.85 & 80.00 & 36.36 \\ 
                APTM (Ours) & \textbf{ 67.50} & \textbf{85.70} & \textbf{91.45} & \textbf{52.56}\\
                \shline
            \end{tabular}
        }
        \vspace{-0.3cm}
        \caption{Performance Comparison on RSTPReid.} 
        \vspace{0.4cm}
        \label{tab:sota_RSTP}
    \end{minipage}
    
    \begin{minipage}[t]{1.0\linewidth}
        \vspace{-0.15cm} 
        \centering
        \small
        \makeatletter\def\@captype{table}\makeatother
        \centering
        \resizebox{1\linewidth}{!}{
            \renewcommand\arraystretch{1} 
            \begin{tabular}[h!]{p{3cm}|m{1.2cm}<{\centering}m{1.2cm}<{\centering}m{1.2cm}<{\centering}m{1.2cm}<{\centering}}
                \shline
                \textbf{Method} & \textbf{R1} & \textbf{R5} & \textbf{R10} & \textbf{mAP}  \\
                \hline
                Dual Path~\cite{zheng2020dual}   & 38.99 & 59.44 & 68.41 & - \\
                CMPM+CMPC~\cite{zhang2018deep}   & 43.51 & 65.44 & 74.26 & - \\
                MIA~\cite{niu2020improving}      & 46.49 & 67.14 & 75.18 & - \\
                SCAN~\cite{lee2018stacked}       & 50.05 & 69.65 & 77.21 & - \\
                ViTAA~\cite{wang2020vitaa}       & 50.98 & 68.79 & 75.78 & - \\
                SSAN~\cite{ding2021semantically} & 54.23 & 72.63 & 79.53 & - \\
                IVT~\cite{shu2023see}            & 56.04 & 73.60 & 80.22 & - \\
                LGUR~\cite{shao2022learning}     & 59.02 & 75.32 & 81.56 & - \\
                CFine~\cite{yan2022clip}         & 60.83 & 76.55 & 82.42 & - \\
                \hline
                Baseline & 57.49 & 75.84 & 82.60 & 32.41 \\ 
                APTM (Ours) & \textbf{68.51} & \textbf{82.99} & \textbf{87.56} & \textbf{41.22} \\
                \shline
            \end{tabular}
        }
        \vspace{-0.3cm}
        \caption{Performance Comparison on ICFG-PEDES.}
        \vspace{0.4cm}
        \label{tab:sota_ICFG} 
    \end{minipage}  
\end{minipage}  
\end{figure*}

\noindent
\textbf{Image-Text Matching Learning}
(ITM) targets to predict whether the input image and the text are matched, analogous to IAM.
Nevertheless, randomly sampling an unpaired item (text or image) is overly facile for the classification. Therefore, we employ a hard example mining strategy.
For each text in a mini-batch, we sample its hard negative image according to the similarity of $S_{\text{t2i}}(T)$, \ie, pick the unpaired image whose similarity is the highest as the hard negative. 
We also sample one hard negative text for each image in a similar manner. 
Finally, $|B|$ positive image-text pairs and $2|B|$ negative pairs, denoted as $\bar{B}$, will pass through the Cross Encoder and one MLP with Sigmoid activation. Following these steps, the ITM loss can be calculated similarly as described in Eq.~\ref{eq:litm}.

\noindent\textbf{Masked Language Modeling}
(MLM) endeavors to predict the masked words using the image and text clue. 
Given an image-text pair $(I, T)$ in $B$, we obtain corresponding masked text $\hat{T}$ following the same masking strategies as MAM. Subsequently, $(I, \hat{T})$ are passed through the encoders to obtain embeddings. The MLM loss $\mathcal{L}_{mlm}$ is analogously imposed following Eq.~\ref{malmLoss}.
Given the above optimization objectives, the full pre-training loss is formulated as:
$\mathcal{L} = \mathcal{L}_{\text{itc}} + \mathcal{L}_{\text{itm}} + \mathcal{L}_{\text{mlm}} + \mathcal \beta \mathcal{L}_{\text{APL}}$,
where $\beta$ denotes the APL loss weight, and we empirically set $0.8$.

\section{Experiment}

\subsection{Experimental Setup}

\textbf{Datasets.} 
We evaluate our approach on three public text-based person retrieval datasets and one  pedestrian attribute dataset,
\emph{i.e.,} \textbf{CUHK-PEDES}~\cite{li2017person},  \textbf{RSTPReid}~\cite{zhu2021dssl}, \textbf{ICFG-PEDES}~\cite{ding2021semantically} and \textbf{PA-100K}~\cite{liu2017hydraplus}. 
In particular, CUHK-PEDES~\cite{li2017person} includes $80,440$ description phrases and $40,206$ photos of $13,003$ people, while RSTPReid~\cite{zhu2021dssl} comprises $20,505$ images of $4,101$ people and is created by compiling MSMT17~\cite{wei2018person} data. 
ICFG-PEDES~\cite{ding2021semantically} is also incubated from MSMT17 and has $54,522$ images of $4,102$ individuals. 
To make a fair comparison, the data splits of three text-based person retrieval datasets keep the same as the previous works~\cite{yan2022clip,farooq2022axm}.
PA-100K~\cite{liu2017hydraplus} is constructed by $100,000$ pedestrian images from $598$ real outdoor scenes.
Every image is labeled by $26$ attributes, and the standard splits are adopted for performance evaluation ~\cite{liu2017hydraplus}.

\noindent\textbf{Implementation Details.}
We pre-train APTM with Pytorch on $4$ NVIDIA A100 GPUs for $32$ epochs, and the mini-batch size is $150$. 
We adopt the AdamW \cite{loshchilov2018decoupled} optimizer with a weight decay of $0.01$. 
The learning rate is decayed from $1e^{-4}$ to $1e^{-5}$ following a linear schedule, after a warm-up schedule beginning at $1e^{-5}$ in the first $2,600$ steps. 
Every image input is resized to $384 \times 128$. 
Random horizontal flipping, RandAugment~\cite{cubuk2020randaugment} and random erasing~\cite{zhong2020random} are employed for image augmentation.
APTM takes text with no more than $56$ tokens as input.
During pre-training, the Image Encoder is initialized with Swin Transformer$_{\text{base}}$~\cite{liu2021Swin}, while the Text Encoder and Cross Encoder are initialized by the first and the last $6$ layers of BERT$_{base}$ \cite{kenton2019bert}, respectively. Therefore, there are $214.5M$ trainable parameters in APTM.
After pre-training, the model is fine-tuned on the downstream datasets for $30$ epochs. 
The learning rate is set as $1e^{-4}$ and is warmed up in the first $3$ epochs.
Then we apply a linear scheduler to gradually decay the learning rate.

\subsection{Comparison with Existing Methods}
We adapt APTM to downstream text-based person retrieval tasks and pedestrian attribute recognition tasks. Following previous practices~\cite{shu2023see,yan2022clip}, we report Recall@1,5,10, and mAP for text-based person retrieval to compare the results. For the attribute recognition task, accuracy (Acc), precision (Prec), recall rate (Rec), and F1 value (F1) are adopted to evaluate the performance.

\noindent\textbf{Text-based Person Retrieval.}
We evaluate APTM on CUHK-PEDES, RSTPReid, and ICFG-PEDES datasets and optimize ITC, ITM, and MLM loss during finetuning. 
Besides image data augmentation mentioned in pretraining, we adopt EDA\cite{wei2019eda} for text data augmentation and set the mini-batch size as $120$.
In reference, for each text query, we first compute its cosine similarity with all images and take the top-$128$ image candidates. Then we calculate the matching probability between the text query and every selected image candidate for ranking.
The proposed method has achieved the SOTA recall rate on all three datasets. Specifically, our model has surpassed $6.96 \%$ recall@1 rate on CUHK-PEDES, compared to the second-best method $69.57 \%$ (\emph{See Table~\ref{tab:sota_CUHK}.}). 
Similarly, as shown in Table~\ref{tab:sota_RSTP} and Table~\ref{tab:sota_ICFG}, we could observe that our method arrives at $67.50 \%$ and $68.51 \%$ R1 on RSTPReid and ICFG-PEDES, respectively.
Furthermore, we compare two traditional methods, TextReID~\cite{han2021text} and IVT~\cite{shu2023see}, with APTM based on MALS. We use ResNet50 as the visual backbone of TextReID and use 30k image-text pairs from MALS for TextReID pretraining since TextReID uses instance loss. IVT also explores the pretraining on 4M general image-text pairs, while MALS only consists of 1.5M image-text pairs.
As shown in Table~\ref{tab:sota_CUHK}, after pretraining on MALS, the performance of all three methods has improved.

\begin{figure}
\vspace{-2.6cm}
\centering
    \begin{minipage}[t]{1.0\linewidth}
        \vspace{-1.3cm} 
        \centering
        \small        \makeatletter\def\@captype{table}\makeatother
        \centering
        \resizebox{1\linewidth}{!}{
            \renewcommand\arraystretch{1} 
            \begin{tabular}[h!]{p{3cm}|m{1.2cm}<{\centering}m{1.2cm}<{\centering}m{1.2cm}<{\centering}m{1.2cm}<{\centering}m{1.2cm}<{\centering}}
                \shline
                \textbf{Method} & \textbf{mA} & \textbf{Acc} & \textbf{Prec} & \textbf{Rec} & \textbf{F1}  \\
                \hline
                HP-net~\cite{liu2017hydraplus} & 74.21 & 72.19 & 82.97 & 82.09 & 82.53 \\
                strongBaseline~\cite{jia2020rethinking} & 79.38 & 78.56 & \textbf{89.41} & 84.78 & 86.55 \\
                ALM~\cite{tang2019improving} & 80.68 & 77.08 & 84.21 & \textbf{88.84} & 86.46\\
                RethinkPAR~\cite{jia2021rethinking} & 81.61 & 79.45 & 87.66 & 87.59 & 87.62 \\
                \hline
                Baseline (Image only) & 71.68 & 54.51 & 60.47 & 83.60 & 70.18 \\
                Baseline (\emph{wo} MAM) & 80.43 & 79.91 & 89.26 & 86.49 & 87.85 \\
                Baseline & 81.49 & 79.89 & 88.59 & 87.09 & 87.83 \\
                APTM (Ours) & \textbf{82.58} & \textbf{80.17} & 88.31 & 87.84 & \textbf{88.07} \\
                \shline
            \end{tabular}
        }
        \vspace{-0.25cm}
        \caption{Performance Comparison on PA-100K. Baseline (Image only) denotes only finetuning the Image Encoder, while Baseline (\emph{wo} MAM) removes MAM loss. Baseline refers to training the model without pretraining on MALS.}
        \vspace{-0.3cm}
        \label{tab:PA} 
    \end{minipage}  
\end{figure}

\begin{figure*}[t]\centering
\vspace{-1em}
\centering\subfloat[Ablation study on the pre-training data number.
\label{fig:scale}
]{\begin{minipage}[b]{0.6\linewidth}
\includegraphics[width=0.83\linewidth]{images/pre_scale.pdf}
\end{minipage}
}
\hspace{-1em}
\subfloat[Ablation study on attribute-oriented objectives.\label{tab:cmal}]{
\begin{minipage}[b]{0.32\linewidth}
\tablestyle{2pt}{1.1}
\begin{tabular}{l|cccc|c}
\shline
Method & $L_{TML}$ & $L_{V1}$ & $L_{V2}$ & $L_{APL}$ & R1 \\
\hline
M1   & $\checkmark$ & & & & 69.62 \\
M2   & $\checkmark$ & $\checkmark$ & & & 69.46 \\
M3   & $\checkmark$ & & $\checkmark$ & & 69.64 \\
APTM & $\checkmark$ & & & $\checkmark$ & 71.38 \\
\shline
\end{tabular}
~\\~\\~\\
\end{minipage}}
\vspace{-.5em}
\caption{Ablation Study on the pre-training data scale and the optimization objectives in our APTM. (a) We apply $0M$, $0.3M$, $0.6M$, $0.9M$, $1.2M$, and $1.5M$ data pairs to pre-train, and then report the fine-tuned recall rate on three datasets respectively. We can observe that the performance is consistently improved as the data scale increases. (b) $L_{TML}$ refers to the sum of ITC loss, ITM loss, and MLM loss, and $L_{APL}$ denotes APL loss.
$L_{V1}$ and $L_{V2}$ are losses of two naïve CMAL variants.
}
\label{ablation}
\end{figure*}

\begin{figure}[t]
\begin{center}
     \includegraphics[width=1.0\linewidth]{images/result.pdf}
\end{center}
\vspace{-.15in}
      \caption{Qualitative text-to-image retrieval results of APTM and baseline, placing in descending order from right to left based on similarity.
      The green boxes indicate the correct matches, and the images in the red boxes are the wrong matches. The green texts highlight the details that our results successfully match.
      }\label{fig:result}
\vspace{-0.5em}
\end{figure}

\noindent\textbf{Pedestrian Attribute Recognition.}
Pedestrian attribute recognition aims at mining the attributes of target people when given a pedestrian image. We apply the attribute prompt learning part of APTM to predict the attributes of images from PA-100K. Similar to MALS, we construct the attribute prompts for PA-100K and finetune the model.
During inference, we compute the matching probability between every image and every pair of attribute prompts for ranking.
An attribute prompt with a higher matching probability means the image is more relevant to the corresponding attribute.
Our method achieves competitive results as shown in Table~\ref{tab:PA}.
Baseline refers to training the model without pretraining on MALS.
Baseline (Image only) denotes only finetuning the Image Encoder and its following MLPs, which are used to predict the attribute label. 
Baseline (\emph{wo} MAM) does not optimize MAM loss.
The results among the three baselines and APTM indicates the rationality of our APTM.
APTM obtain $0.97 \%$ improvement on mA compared with the results of RethinkPAR~\cite{jia2021rethinking}. 
A recent work SOLIDER~\cite{chen2023beyond} reports $86.37 \%$ mA. Since SOLIDER~\cite{chen2023beyond} adopts a more powerful backbone, we do not include the result of SOLIDER for a fair comparison.

\subsection{Ablation Study}

\noindent\textbf{Effectiveness of Pre-Training.} 
Table~\ref{tab:sota_CUHK} compares the performance on the CUHK-PEDES dataset, where the "Baseline" means the APTM without pre-training. We can observe that our baseline, reaching $66.44 \%$, $84.92 \%$, and $90.76 \%$ on R1, R5, and R10, respectively, is a competitive method. Pre-training APTM on MALS leads to improvement of $10.09 \%$, $5.12 \%$, and $3.39 \%$ on Recall@1, 5 and 10. Similar results could be observed on RSTPReid and ICFG-PEDES, reported in Table~\ref{tab:sota_RSTP} and Table~\ref{tab:sota_ICFG}. 
To intuitively show the benefits of pre-training, three qualitative results of APTM and baseline are shown in Figure~\ref{fig:result}, indicating the superiority of APTM.

\noindent\textbf{The Impact of Pre-training Scale.} 
In generic vision and language pre-training tasks, the scale of the training dataset usually plays an important role. A larger amount of pre-training data often means better performance.
To thoroughly study the effectiveness of MALS, we further explore the impact of the data scale during pre-training. Specifically, we respectively adopt $0$, $0.3M$, $0.6M$, $0.9M$, $1.2M$ and, $1.5M$ data of MALS to pre-train 32 epochs and then evaluate the finetuned performance on CUHK-PEDES, ICFG-PEDES and RSTPReid.
The results are compared in Figure~\ref{fig:scale}, as the data scale increases, the recall tends to improve as well. From $0$ to $0.3M$, finetuning performance on three datasets increases noticeably, while from $0.3M$ to $1.5M$, the rate of improvement gradually diminishes.

\noindent\textbf{Effectiveness of APL Loss.} 
We also conduct an ablation study to investigate how to leverage the attribute annotation, as shown in Table~\ref{tab:cmal}. 
All compared model variants are pre-trained on $0.03M$ data of MALS and then finetuned on CUHK-PEDES.
We adopt Recall@1 as an evaluation measure.
First, we evaluate the effectiveness of APL loss, \ie, M1, APTM. 
The results show that pretraining without APL loss hurts performance.
Furthermore, we replace APL with several naïve CMAL variants separately, and report its finetuning performance on CUHK-PEDES in Table~\ref{tab:cmal}:
(1)Method V1: Image embedding and Text Embedding are used to give the prediction of attributes by mapping the Embedding to low-dimensional features, separately. The BCE Loss is adopted as the objective function.
(2)Method V2: Use the joint representation of the image-text pair to predict attributes by mapping the Embedding to low-dimensional features. The attribute classification loss is BCE Loss, too. 
In Table~\ref{tab:cmal}, compared with M2 and M3, APL
outperforms both of them, which verifies the superiority of APTM.

\section{Conclusion}
We introduce MALS, a new large-scale benchmark for multi-attribute recognition and language-based person search. Our benchmark comprises $1,510,330$ image-text pairs with rich attribute annotations, which is about 37.5 times larger than widely-used CUHK-PEDES. Extensive experiments verify that pretraining on MALS is scalable to real-world scenarios. To regularize the model training, we propose to jointly learn from the two complementary tasks, \ie, text-based person retrieval and pedestrian attribute recognition.
On three public benchmarks, including CUHK-PEDES, ICFG-PEDES, and RSTPReid, our approach has achieved a competitive recall rate.
We hope our work could contribute to the community with a new viewpoint on unified text-based person retrieval.

\noindent\textbf{Broader Impact.}
The proposed MALS could facilitate the person retrieval task on limited data, and help the community to have a large-scale dataset for pre-training. 
(1) Language-based person search is a more intuitive way for users for city safety, \eg, finding lost children at airports or parks.
(2) Since our dataset is generated, we do not have access to any specific person, meeting privacy concerns. The generation process also largely saves the tedious manual annotation.

\noindent\textbf{Acknowledgments}
This work is supported by the National Key Research and Development Project, China (No. 2019YFB2102501).

\definecolor{gain}{HTML}{77ac30}  
\newcommand{\gain}[1]{\textcolor{gain}{#1}}
\definecolor{lost}{HTML}{ea4335}  
\newcommand{\lost}[1]{\textcolor{lost}{#1}}
\newcommand{\res}[2]{{#1} {({\gain{#2}})}}

{\small
\bibliographystyle{ACM-Reference-Format}
\bibliography{egbib}
}

\clearpage
\appendix
{\sc{\huge{Supplementary Material}}}

\section{Network Details}
APTM consists of the Image Encoder $E_I$, Text Encoder $E_T$, Cross Encoder $E_C$, and two MLPs-based headers. As described in the paper, $E_I$ is Swin-B, while $E_T$ and $E_C$ are the first $6$ layers and the last $6$ layers of Bert, respectively. 
Here we show the parameters and GFLOPs of $E_I$, $E_T$, $E_C$, and the whole APTM in Table~\ref{table:params}. 
The inference time per text query of APTM is $4.8ms$.

\section{Attribute Prompt Template}
In Attribute Prompt Learning, we map 27 binary attributes to 54 Attribute Prompts and then align the attribute prompts with the corresponding image. 
Inspired by the “prompt engineering” discussion in GPT3~\cite{brown2020GPT3, gao2020making} and CLIP~\cite{radford2021CLIP}, we convert attribute labels into attribute prompts with prompt templates.
Specifically, we customize different prompt templates for different attributes as shown in Table~\ref{table:prompts}, while Figure~\ref{fig: framework} in the paper shows some examples of attribute prompts. 
We design five kinds of templates, \ie, "\textit{the person is \{ Label Text \}}", "\textit{the person with \{ Label Text \}}", \emph{etc}.
In Table \ref{table:Attribute}, "Age" is a quaternary attribute, \ie, "young", "teenager", "adult" and "old", however, in MALS, there is a lack of data for "young" and "old". Therefore, in this paper, we treat age as a binary attribute, \ie, "young" and "adult".
We find that using the attribute prompt templates in Table~\ref{table:prompts} could be a good default, which often improves performance over using only the label text. 
The templates specify the attribute prompt is about the attribute content of the image and bridge the distribution gap between attribute prompts and original image paired text.

\section{Further Experiments and Discussion}
\textbf{Parameter sensitivity of $\beta$.} 
We further pre-train APTM on 0.03M data of MALS for $32$ epochs and compare the impact of different values of $\beta$ ($\beta$ is the weight of APL loss) in Equation 8 on model performance. 
We set $\beta$ as $0, 0.4, 0.6, 0.8, 1.0, 1.4$ and $2.0$, respectively during pre-training and conduct the same fine-tuning on the CUHK-PEDES dataset. 
As shown in Table~\ref{table:beta}, it could be observed that the model with $\beta = 0.8$ achieves best performance (Recall@1 (R1) $= 71.38 \%$).

\noindent\textbf{Robustness against Broken Sentences.} 
Pre-trained APTM shows robustness and generalization on the downstream task, \ie, text-based person retrieval. We finetune the Pre-trained APTM on CUHK-PEDES and exploit the performance of the model by deleting some crucial words in the text query. We test our model by randomly masking some words in a text with a special unknown token \texttt{[UNK]} using a probability of $0.1$. Figure~\ref{fig:roubustness} are four obtained sample results, which demonstrate that APTM still performs well even in the presence of obstacles. In the first row, despite "street" being masked out, APTM remains sensitive to "crossing" and infers the right image. In the second row, the color of the jacket and the item held in the hand are masked out, leading to diverse search results, but the model still identifies the right matching option. In the third row, the model infers "white" corresponds to the upper garment, while in the fourth row, the model infers "black" corresponds to the shoes.

\vspace{0.5cm}
\begin{table}[bp]
\small
\begin{center}
{
\setlength{\tabcolsep}{7pt}
\begin{tabular}{l|cccc}
\shline
Module          & $E_I$ & $E_T$ & $E_C$ & APTM \\
\hline
Parameters & $86.8M$ & $66.4M$ & $59.1M$ & $214.5M$ \\
GFLOPs     & $14.9$ & $2.4$ & $3.2$ & $38.0$ \\
\shline
\end{tabular}}
\end{center}
\caption{The parameters and GFLOPs of $E_I$, $E_T$, $E_C$ and the whole APTM.
}
\label{table:params}
\end{table}

\begin{table}[bp]
\small
\begin{center}
{
\setlength{\tabcolsep}{7pt}
\begin{tabular}{l|ccccccc}
\shline
$\beta$ & 0     & 0.4   & 0.6   & 0.8   & 1.0   & 1.4   & 2.0  \\
\hline
R1      & 69.62 & 70.29 & 70.63 & 71.38 & 71.25 & 70.91 & 70.65 \\
\shline
\end{tabular}}
\end{center}
\caption{ The impact of $\beta$ on APTM. We set $\beta = \{0, 0.4, 0.6, 0.8, 1.0, 1.4, 2.0\}$ respectively and compare the performance of the corresponding pre-trained model by finetuning the model on CUHK-PEDES and Recall@1 (R1) is reported.
}
\label{table:beta}
\end{table}

\begin{figure}[bp]
\begin{center}
   \includegraphics[width=\linewidth]{images/robust.pdf}
\end{center}
\vspace{-.1in}
   \caption{Examples of the text-based person retrieval results, demonstrating the robustness of APTM. The searched images are placed in descending order from left to right based on matching probability. We randomly replace some words in the sentence with a special unknown token \texttt{[UNK]}. The green boxes indicate correct matches, and the images in the red boxes are wrong matches. The green texts highlight the replaced positions. This figure is best viewed when zooming in.}
\label{fig:roubustness}
\end{figure}

\setlength{\tabcolsep}{3pt}
\begin{table*}
\small
\begin{center}
\resizebox{1\linewidth}{!}{\begin{tabular}{l|c|c|c}
\hline
Attribute Prompt Template  &  Attribute Name  &  Label  &  Label Text  \\
\shline

\multirow{2}{*}{\textit{the person is \{ Label Text \}}}  
& gender & female(0), male(1) &  \textit{a woman, a man} \\
& age    & young(0), adult(1) &  \textit{younger than 18 years old, older than 18 years old} \\
\hline

\multirow{3}{*}{\textit{the person with \{ Label Text \}}}
& length hair       & short hair(0), long hair(1) &  \textit{short hair, long hair} \\
& wearing hat                     & yes(0)  &  \textit{a hat} \\
& carrying backpack, handbag, bag & yes(0)  &  \textit{a backpack, handbag, bag} \\
\hline

\multirow{2}{*}{\textit{the person without \{ Label Text \}}}
& wearing hat                     & no(1)  &  \textit{a hat} \\
& carrying backpack, handbag, bag & no(1)  &  \textit{a backpack, handbag, bag} \\
\hline

\multirow{7}{*}{\textit{the person wears \{ Label Text \}}}
& sleeve length        & long sleeve(0), short sleeve(1)       & \textit{long sleeved upper clothes, short sleeved upper clothes}  \\
& length of lower-body & long lower-body clothing(0), short(1) & \textit{long dress or long pants, short dress or short pants}  \\
& type of lower-body   & dress(0), pants(1)                    & \textit{dress or skirt, pants or shorts}  \\

& upper-body black, white, red, purple,    & \multirow{4}{*}{yes(0)} & \textit{black, white, red, purple,} \\
& yellow, blue, green, gray                &                         & \textit{yellow, blue, green, gray upper clothes} \\
& lower-body black, white, purple, yellow, &                         & \textit{black, white, purple, yellow,} \\
& blue, green, pink, gray, brown           &                         & \textit{blue, green, pink, gray, brown lower clothes} \\
\hline

\multirow{4}{*}{\textit{the person does not wear \{ Label Text \}}}
& upper-body black, white, red, purple,    & \multirow{4}{*}{no(1)}  & \textit{black, white, red, purple,} \\
& yellow, blue, green, gray                &                         & \textit{yellow, blue, green, gray upper clothes} \\
& lower-body black, white, purple, yellow, &                         & \textit{black, white, purple, yellow,} \\
& blue, green, pink, gray, brown           &                         & \textit{blue, green, pink, gray, brown lower clothes} \\
\hline

\end{tabular}}
\end{center}
\caption{Five prompt templates of 27 attributes and different label texts for different attribute labels. A whole attribute prompt consists of a prompt template and corresponding label text. 
}
\label{table:prompts}
\end{table*}

\end{document}